\documentclass[letterpaper, 10 pt, conference]{ieeeconf}  

\IEEEoverridecommandlockouts                              

\usepackage{amsmath} 
\usepackage{amssymb}  
\usepackage{graphicx}
\usepackage{url}
\usepackage{multirow}
\usepackage{xcolor}
\usepackage{capt-of}

\usepackage{microtype}
\usepackage{hyperref}
\usepackage{url}
\usepackage{booktabs}

\usepackage{float}
\usepackage{listings}
\usepackage[breakable,skins]{tcolorbox}

\usepackage{lineno}

\definecolor{darkblue}{rgb}{0, 0, 0.5}
\hypersetup{colorlinks=true, citecolor=darkblue, linkcolor=darkblue, urlcolor=darkblue}

\lstdefinelanguage{json}{
    basicstyle=\normalfont\ttfamily,
    breaklines=true,
    backgroundcolor=\color{gray!10},
    showstringspaces=false,
    string=[db]{"},
    stringstyle=\color{green!50!black},
    morestring=[s][\color{black}]{\ \ "}{":},
    keywordstyle=\color{blue},
    keywords={true,false,null},
    literate=
     *{0}{{{\color{red}0}}}{1}
      {1}{{{\color{red}1}}}{1}
      {2}{{{\color{red}2}}}{1}
      {3}{{{\color{red}3}}}{1}
      {4}{{{\color{red}4}}}{1}
      {5}{{{\color{red}5}}}{1}
      {6}{{{\color{red}6}}}{1}
      {7}{{{\color{red}7}}}{1}
      {8}{{{\color{red}8}}}{1}
      {9}{{{\color{red}9}}}{1}
      {.}{{{\color{red}.}}}{1}
      {:}{{{\color{gray}{:}}}}{1}
      {,}{{{\color{gray}{,}}}}{1}
      {\{}{{{\color{gray}{\{}}}}{1}
      {\}}{{{\color{gray}{\}}}}}{1}
      {[}{{{\color{gray}{[}}}}{1}
      {]}{{{\color{gray}{]}}}}{1},
}

\definecolor{assistantonecolor}{RGB}{19,118,188}
\definecolor{assistanttwocolor}{RGB}{229,91,43}
\newcommand{\assistantonemsg}[1]{\textcolor{assistantonecolor}{{\textbf{#1: }}}}
\newcommand{\assistanttwomsg}[1]{\textcolor{assistanttwocolor}{{\textbf{#1: }}}}
\tcbset{
  aiboxbreakable/.style={
    width=400.18663pt,
    top=10pt,
    colback=black!05,
    colframe=black!20,
    colbacktitle=black!50,
    enhanced,
    center,
    breakable,
    attach boxed title to top left={yshift=-0.1in,xshift=0.15in},
    boxed title style={boxrule=0pt,colframe=white,},
  }
}
\newtcolorbox{AIBoxBreak}[2][]{aiboxbreakable,title=#2,#1}

\title{\LARGE \bf
Multimodal Reference Visual Grounding}

\author{Anonymous Author(s)}
\author{Yangxiao Lu, Ruosen Li, Liqiang Jing, Jikai Wang, \\
Xinya Du, Yunhui Guo, Nicholas Ruozzi, Yu Xiang
\\
Department of Computer
Science\\
University of Texas at Dallas\\
Richardson, TX 75080, USA \\
\small{\texttt{\{yangxiao.lu, ruosen.li, liqiang.jing, jikai.wang,}}\\
\small{\texttt{xinya.du, yunhui.guo, nicholas.ruozzi, yu.xiang\}@utdallas.edu}}
}
\begin{document}

\makeatletter
\let\@oldmaketitle\@maketitle
\renewcommand{\@maketitle}{\@oldmaketitle
\centering
\vspace{1em}
\includegraphics[width=0.85\linewidth]{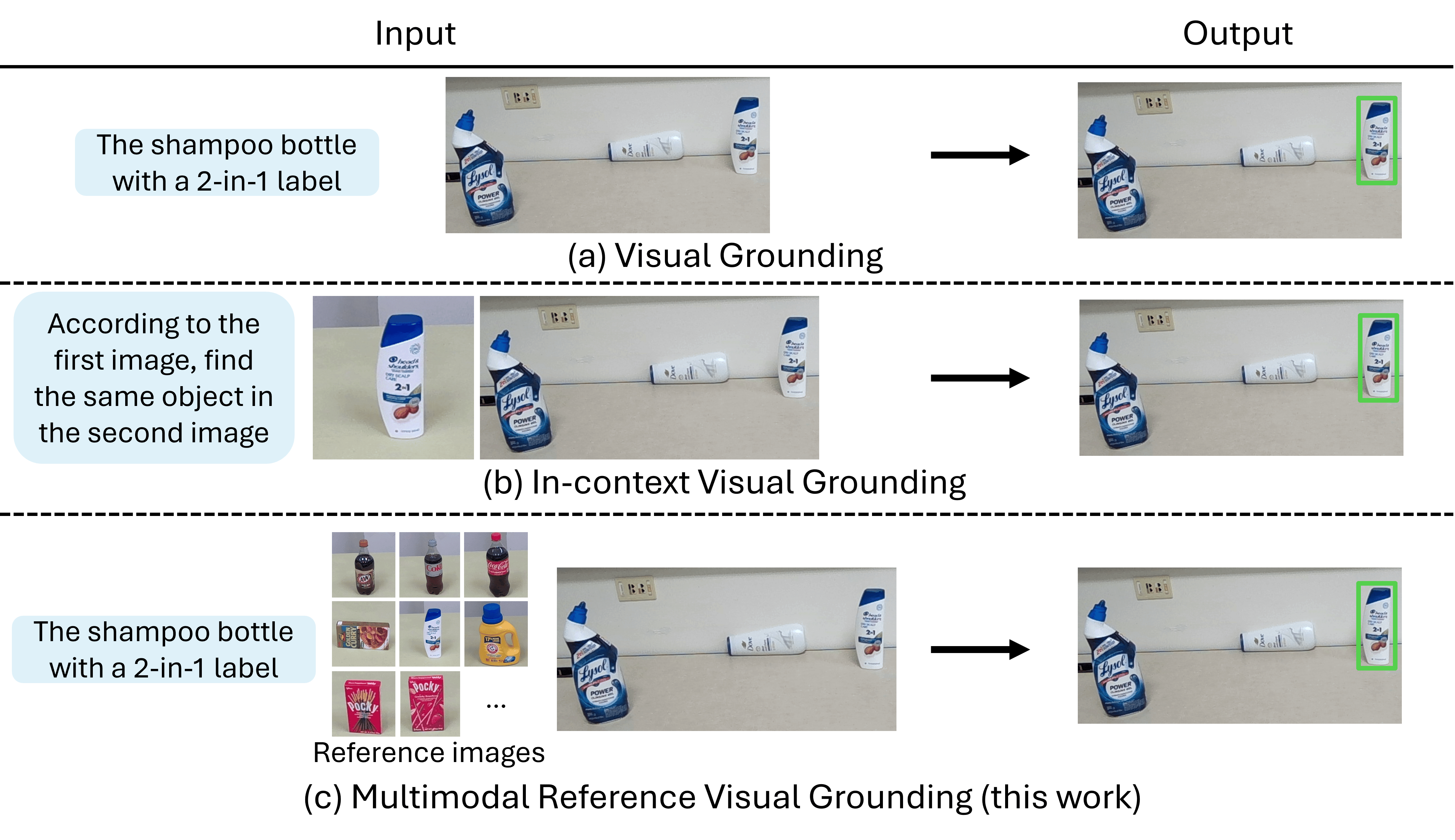}
\vspace{-1.0em}
\captionof{figure}{Comparison of three visual grounding tasks: (a) Visual grounding identifies objects in a query image based on a textual expression. (b) In-context visual grounding utilizes reference images to specify the target object in addition to the language expression, where the reference images must contain the target object. (c) Multimodal reference visual grounding uses a set of reference images alongside a referring expression to identify the target, where the target only represents one object in the reference images.}
\label{fig:grounding}
}
\makeatother

\maketitle
\thispagestyle{empty}
\pagestyle{empty}

\begin{abstract}


Visual grounding focuses on detecting objects from images based on language expressions. Recent Large Vision-Language Models (LVLMs) have significantly advanced visual grounding performance by training large models with large-scale datasets. However, the problem remains challenging, especially when similar objects appear in the input image. For example, an LVLM may not be able to differentiate Diet Coke and regular Coke in an image. In this case, if additional reference images of Diet Coke and regular Coke are available, it can help the visual grounding of similar objects.

In this work, we introduce a new task named Multimodal Reference Visual Grounding (MRVG). In this task, a model has access to a set of reference images of objects in a database. Based on these reference images and a language expression, the model is required to detect a target object from a query image. We first introduce a new dataset to study the MRVG problem. Then we introduce a novel method, named MRVG-Net, to solve this visual grounding problem. We show that by efficiently using reference images with few-shot object detection and using Large Language Models (LLMs) for object matching, our method achieves superior visual grounding performance compared to the state-of-the-art LVLMs such as Qwen2.5-VL-72B. Our approach bridges the gap between few-shot detection and visual grounding, unlocking new capabilities for visual understanding, which has wide applications in robotics\footnote{Project page with our video, code and dataset: \url{https://irvlutd.github.io/MultiGrounding}}.

\end{abstract}

\section{INTRODUCTION}


In the visual grounding task, the goal is to localize an object in a given query image that matches a given textual expression (see Fig.~\ref{fig:grounding}(a)). As an example application, in the robotics domain, a user might instruct a robot bartender to ``bring me a Diet Coke''. To correctly complete the task, the robot bartender must open the refrigerator and find a Diet Coke among many possibly similar items, e.g., regular Coke, zero Coke, or other soft drinks. More generally, to be a competent bartender, the robot must be able to accurately distinguish between many similar types of beverages and ingredients required in cocktail recipes.

Recent large vision-language models (LVLM) ~\cite{wang2023one, bai2025qwen2,chen2024expanding,wang2024cogvlm} have demonstrated impressive performance on visual grounding tasks. However, in settings with very similar objects, such as in the robotics scenario described above, these LVLMs can struggle to differentiate them, which indicates that the LVLMs may not pay attention to the fine-grained details necessary to correctly identify the target object. 


To overcome this limitation, one idea is to incorporate visual references into the query: For example, we can download images of Diet Coke and regular Coke from the Internet and provide these reference images along with the textual expression for visual grounding. In this way, the model should be able to detect a target object based on a referring expression and the reference images of objects. However, many LVLMs~\cite{liu2024grounding,bai2025qwen2,chen2024expanding,wang2024cogvlm} operate in a zero-shot manner. They generate bounding boxes solely on the basis of the provided referring expression. These models do not support the use of reference images to aid inference. As a result, while they may be effective in distinguishing between different object categories, their ability to differentiate highly similar object instances is often quite limited.

To incorporate visual references, recent LVLMs such as DeepSeek-VL2~\cite{wu2024deepseek} enable one-shot in-context visual grounding: Given a reference image and an optional text prompt, DeepSeek-VL2 identifies a target object in a query image. The in-context visual grounding task is illustrated in Fig.~\ref{fig:grounding}(b). However, their approach assumes that the target object must be present in the reference image. When provided with a large number of reference images containing potential target objects, DeepSeek-VL2 struggles to perform in-context visual grounding if only a subset of these targets appears in the query image. This can be a significant limitation in practice.  Consider again the example of a robot bartender.  It is reasonable to assume that the robot possesses visual references for all of the items it has in stock.  As such, it would be convenient to be able to pass its entire stock, which could include thousands of items, as visual references for any visual grounding task.

In this work, we propose the Multimodal Reference Visual Grounding (MRVG) task, which requires a model to generate bounding boxes or segmentation masks of target objects based on textual queries and multiple reference images. The MRVG task is illustrated in Fig.~\ref{fig:grounding}(c). The textual query can be explicitly or implicitly relevant to the target object such as \emph{``A bottle of black liquid with a gray cap and a diet cola label on its side''} or \emph{``The leftmost bottle''}. The reference images contain all potential objects that are relevant to an underlying application. For example, we can use the images of objects in an inventory list in a grocery store as reference images. Or if a robot needs to recognize a set of objects, we can capture images of these objects and use them as the reference images for visual grounding. To address this task, the model must possess the capability to process reference images efficiently to extract visual cues and comprehend the associated query texts.

Second, since there is no existing dataset designed for the MRVG task, we constructed a dataset consisting of 1,400 reference images, 250 query images, and 855 associated referring expressions for 100 everyday objects across 4 different scenes, both indoor and outdoor. We chose objects that are commonly found in daily life as we anticipate that visual grounding of these objects could be useful in a variety of robotics and VR/AR applications.

Finally, we propose a new method, MRVG-Net, to tackle the multimodal reference visual grounding task. In our method, given reference images of target objects, an LVLM such as GPT-4o-mini~\cite{hurst2024gpt} is used to extract text descriptions such as color, shape, and usage of these objects. This step only needs to be run once. Then we obtain textual descriptions of all the objects in the reference images. Given a query image during inference, a few-shot detector, e.g., NIDS-Net~\cite{lu2024adapting}, is first used to detect target objects from the query image. NIDS-Net uses the reference images as a support set for few-shot object detection. Consequently, it only detects objects that are present in the reference images. Using the predicted object instance IDs from NIDS-Net, we can retrieve the corresponding object descriptions of the detected objects. Finally, an LLM such as GPT-4o~\cite{hurst2024gpt} performs reasoning to match the object descriptions with the input textual query to generate the final object grounding results. 
 Instead of relying on an LVLM to process the reference images as in in-context visual grounding, our method utilizes a few-shot object detector to detect objects based on the reference images. As a result, our method can handle a large number of reference images. In addition, object matching is achieved using text only, where we can leverage the state-of-the-art LLMs for robust matching.


We evaluated MRVG-Net on our new dataset, and we compared it with several visual grounding methods such as GroundingDINO~\cite{liu2024grounding}, InternVL2.5~\cite{chen2024expanding}, and Qwen2.5-VL~\cite{bai2025qwen2}. Our method achieves superior visual grounding performance compared to these methods, which demonstrates the effectiveness of our method in solving the MRVG task and exemplifies the novelty of the task.

In summary, the contributions of this work are as follows.
\begin{itemize}
    \item We introduce the task of multimodal reference visual grounding, where reference images of target objects are available for visual grounding.
    \item We contribute a new dataset for multimodal reference visual grounding that contains 100 everyday objects such as drink bottles and snack boxes.
    \item We introduce a new method for multimodal reference visual grounding that leverages few-shot object detection and LVLM/LLM for object matching.
\end{itemize}

\addtocounter{figure}{-1}
\begin{figure*}
    \centering
\includegraphics[width=0.95\linewidth]{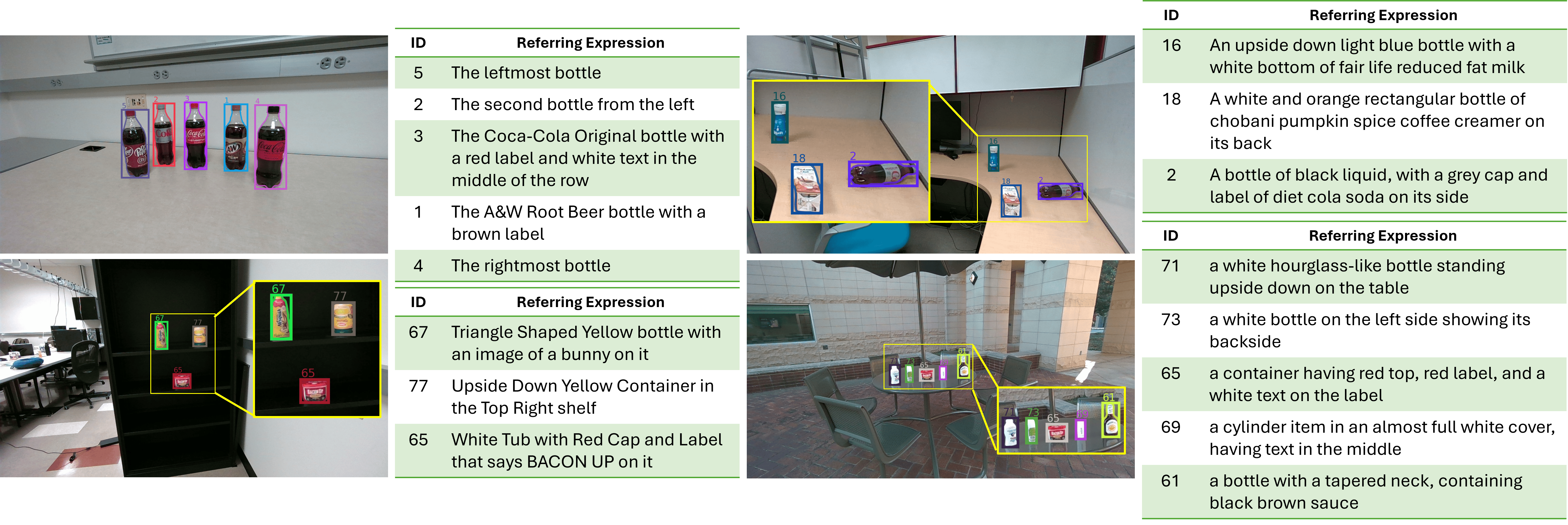}
    \caption{Illustration of the four different scenes in our MultimodalGround dataset. Each target object is annotated with an object instance ID, a bounding box, a segmentation mask, and a referring expression.} 
    \label{fig:scenes}
\end{figure*}

\begin{figure*}
    \centering
\includegraphics[width=0.9\linewidth]{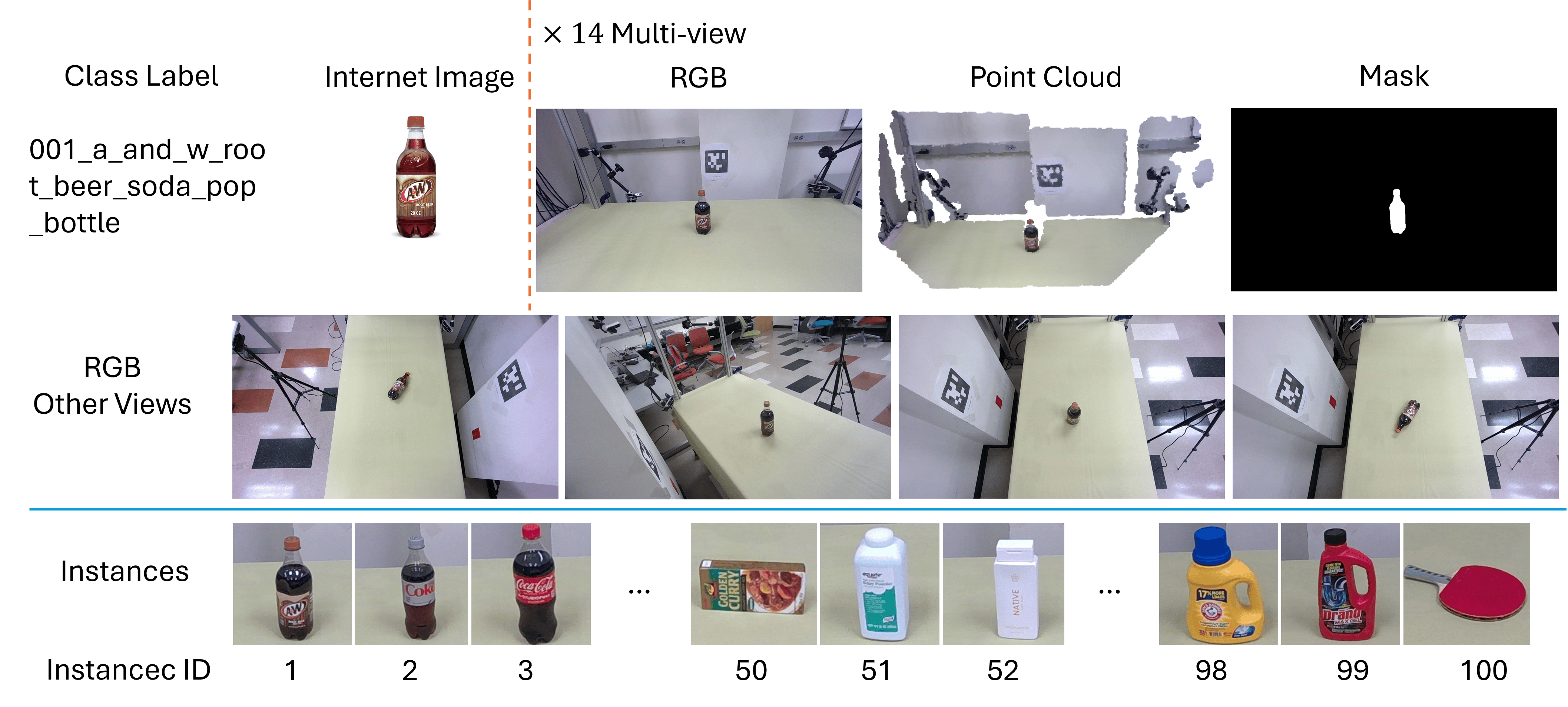}
    \caption{Our MultimodalGround dataset comprises 100 everyday object instances. Each instance is annotated with a class label and instance ID, and is accompanied by one detailed image, 14 multi-view RGB-D reference images, and corresponding segmentation masks. Point clouds can be generated using the provided camera intrinsics and depth images.}
\label{fig:template}
\end{figure*}

\section{Related Work}
\textbf{Visual Grounding (VG)}, also known as Referring Expression Comprehension (REC), is the task of identifying a region in an image according to a query text. This task requires understanding free-form text to locate objects from any open-set category. REC benchmarks, such as RefCOCO, RefCOCO+, and RefCOCOg~\cite{yu2016modeling}, provide a large number of annotations. However, these datasets do not provide reference images that explicitly highlight the objects referred to by the query text within the images. Specialized VG models have been proposed in previous works~\cite{yu2018mattnet, kamath2021mdetr, UNINEXT, wu2022referformer}. Recently, Large Vision-Language Models~\cite{liu2024grounding, chen2024expanding, bai2025qwen2, wu2024deepseek, llava-grounding} have demonstrated significant performance improvements on the traditional VG datasets due to their large training data. These models primarily focus on localizing objects using only the referring expression, without relying on visual cues as references. In contrast, we develop a new framework to utilize reference images as additional contextual information.

\textbf{Novel instance detection} identifies and labels unseen object instances in a query image using template/reference images. VoxDet~\cite{li2024voxdet} leverages multi-view templates to provide geometry-invariant 3D representations. \cite{shen2023instance} introduces a high-resolution dataset and utilizes SAM~\cite{kirillov2023segment} and DINOv2~\cite{oquab2023dinov2} to detect instances. NIDS-Net~\cite{lu2024adapting} uses Grounded-SAM~\cite{ren2024grounded} for object proposals and refines DINOv2 embeddings with a weight adapter. We adopted NIDS-Net in our method on the basis of its strong practical performance.

\textbf{LLM Reasoning} capabilities can be substantially enhanced during inference using various test-time prompting techniques. Approaches such as Chain of Thought \cite{wei2022chain} have been shown to significantly improve reasoning performance. Additionally, Process Reward Models (PRMs) can effectively guide the inference process through diverse strategies such as Majority Voting \cite{wang2022self}, Tree-of-Thought \cite{yao2023tree}, Beam Search \cite{lewis2019bart}, and Lookahead Search \cite{snell2024scaling}. We leverage the reasoning ability of LLMs to better understand the relationships between item descriptions and user queries.

\textbf{Large Vision-Language Model.} 
LVLMs have received significant attention for their impressive visual-understanding capabilities. For instance, LLaVA \cite{DBLP:conf/nips/LiuLWL23a} integrates a visual encoder with a large language model to enable vision-centric multi-turn dialogue. Most existing LVLMs \cite{MultiModal-GPT,mPLUG-Owl,fgaif,InstructBLIP} adopt a common architecture that includes a visual encoder, a language decoder, and a projection layer to align modalities. Although these models have achieved remarkable success across a wide range of visual understanding tasks, their ability to perform grounded understanding, i.e., precisely predicting bounding boxes based on the language instruction, remains limited.
To address this limitation, LLaVA-Grounding \cite{llava-grounding} introduces a new grounded visual dialogue dataset to enhance grounding performance during visual conversations. In contrast to existing grounding-capable LVLMs such as DeepSeek-VL2 \cite{deepseekvl2} and Qwen2.5-VL \cite{qwenvl25}, our work uniquely focuses on how to predict bounding boxes of objects in a scene image with both textual natural language instructions and reference images. 
\section{Multimodal Reference Visual Grounding}
Visual grounding tasks traditionally identify and localize objects within an image based on a referring expression. In-context visual grounding enhances this by using one or more reference images to more accurately locate the target object within the query image. However, this method assumes the presence of the target object in the reference images, which is not always practical. As illustrated in Fig.~\ref{fig:grounding}, we propose a new task, termed Multimodal Reference Visual Grounding, which expands the scope from using a limited number of relevant reference images to potentially a large number of reference images. The model must autonomously select pertinent reference images to effectively gather visual context for language grounding.


Specifically, in multimodal reference visual grounding, we are provided with a set of \( N \) target instances, where each instance is represented by \( K \) reference images. We denote all the reference images as \( I_T \in \mathbb{R}^{N \times K \times 3 \times W \times H} \), where $W$ and $H$ denote the width and height of the images, respectively. Each reference image contains only one object with a segmentation mask of the object. We denote all segmentation masks of the reference images as \( M_T \in \mathbb{R}^{N \times K \times W \times H} \). The objective is to locate and segment a specific target object described by a referring expression \( r \) within a query image \( I_Q \in \mathbb{R}^{3 \times W \times H} \). The referring expression \( r \) provides a natural language description that unambiguously identifies one of the \( N \) target objects that appears in the query image. Each of the \( K \) template images associated with a target object serves as a visual reference, which the model utilizes to match the referred object in the query image.
The output consists of a bounding box \( b \) and an optional instance segmentation mask \( m \in \mathbb{R}^{W \times H} \) that precisely delineates the boundary of the object.

\begin{figure*}
    \centering
\includegraphics[width=0.9\linewidth]{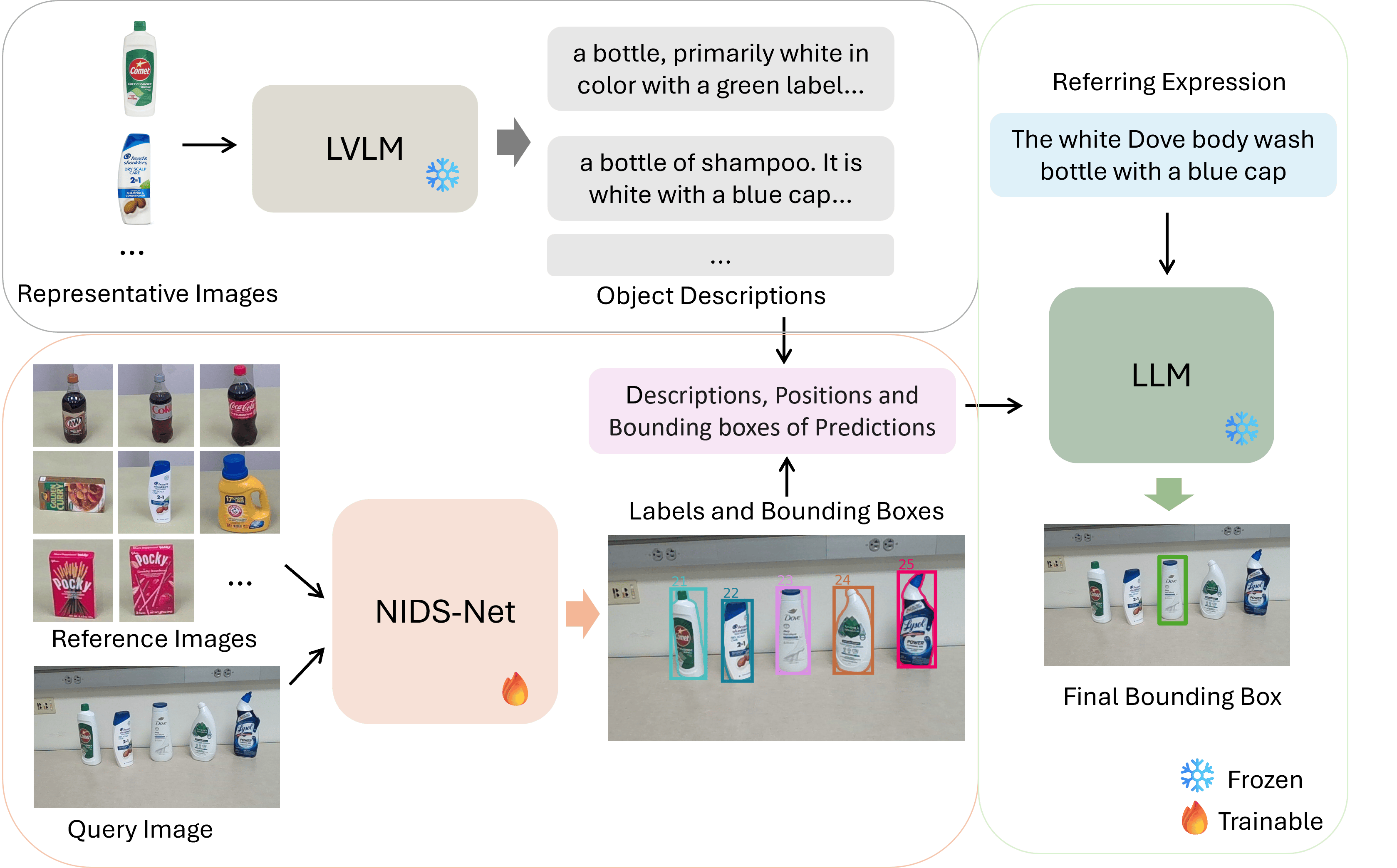}
    \caption{Architecture of our MRVG-Net. Only the weight adapter of NIDS-Net~\cite{lu2024adapting} is trained using the reference images. After the NIDS-Net produces detection results in a few-shot fashion, the corresponding object descriptions are retrieved from a pre-stored file. The Large Language Model (LLM) then matches these descriptions with the referring expression to select the most relevant description and its associated bounding box.}
    \label{fig:framework}
    \vspace{-4mm}
\end{figure*}

\subsection{The MultimodalGround Dataset}

We introduce a new dataset called \textit{MultimodalGround} which can be used to evaluate models on two tasks: standard visual grounding and multimodal reference visual grounding. The dataset consists of 250 RGB-D query images and 855 object annotations manually labeled by humans. These query images are used only for \textbf{evaluation}, not for training. These images are captured using one Intel RealSense D455 camera. Each annotation includes a bounding box, a segmentation mask, and a referring expression as shown in Fig. \ref{fig:scenes}. The average lengths of query texts in RefCOCO, RefCOCO+, and RefCOCOg~\cite{yu2016modeling} are 3.6, 3.6, and 8.4 words, respectively, while the average referring expression length in our dataset is 9.4 words to include specific object details. When constructing these annotations and query images, we imagine that a robot or an AI agent is instructed to locate specific objects and engage in further interactions, such as grasping and manipulation in real-world scenarios. Hence, these annotations can be used to assess the real-world performance of various models. 

The objects in the query images represent 100 common daily items that can be found in supermarkets. For each object instance, the dataset provides 14 real-world RGB-D template images captured from seven viewpoints using seven Intel RealSense D455 cameras, along with one RGB image sourced from the Internet to support the generation of detailed object descriptions. For each RGB-D reference image, we provide a corresponding mask for each object, as illustrated in Fig. \ref{fig:template}. These reference images are used to provide visual references for language grounding and can be used for training. To facilitate future research in RGB-D visual grounding, novel instance detection and segmentation, we provide both camera intrinsic parameters and depth images in our dataset.

We focus on supermarket objects, where identifying visually similar items presents a realistic and challenging task for robotic perception. To our knowledge, this is the first benchmark to define multimodal reference visual grounding at the instance level, supporting fine-grained understanding and future research in broader domains.

\section{Method}

Our framework MRVG-Net adopts a detection-and-matching paradigm, as illustrated in Fig.~\ref{fig:framework}. For each of the $N$ objects, a representative image $I_i$ is used to generate a detailed object profile $D_i$ using a Large Vision-Language Model (LVLM), where $i = 1, \ldots, N$. 
These profiles $\{D_i\}_{i=1}^N$ serve as the basis for object matching. 
Given a query image $I_Q$, the few-shot detector NIDS-Net~\cite{lu2024adapting} identifies relevant objects by producing a set of object instance IDs $\{c_j\}$ and their corresponding bounding boxes $\{b_j\}$, where $j = 1, \ldots, M$ for $M$ detections, based on the reference images.
Using the predicted object instance IDs, we retrieve the corresponding object descriptions $\{D_{c_j}\}$ and spatial positions $\{b_j\}$ to form a set of candidate objects $\mathcal{C} = \{(D_{c_j}, b_j)\}_{j=1}^M$. 
Finally, a LLM performs matching between the candidates $\mathcal{C}$ and the referring expression $R$ to identify the target object.

\subsection{Object Descriptions}


To support accurate object grounding, we employ a Large Vision-Language Model (LVLM) to generate detailed natural language descriptions for $N = 100$ daily objects, each represented by a high-quality Internet image $I_i$, where $i = 1, \ldots, N$. For each object, the LVLM produces a descriptive profile $D_i$ that captures fine-grained visual attributes such as shape, color, material, and function. These descriptions $\{D_i\}_{i=1}^N$ are context-aware, align well with natural referring expressions, and enhance cross-modal retrieval performance. The object descriptions can be precomputed using an LVLM and stored in a file, such that the LVLM is not utilized during inference. Detailed LVLM prompts and generation examples are provided in the appendix~\ref{sec:app_object}. 

\subsection{Novel Instance Detection}

Given reference images $I_T$ of the target instances, NIDS-Net~\cite{lu2024adapting} detects corresponding objects in a query image $I_Q$ through the following steps. First, it applies Grounding DINO~\cite{liu2024grounding} to $I_Q$ using a generic text prompt (e.g., ``objects'') to obtain a set of initial bounding boxes $\{b_j\}_{j=1}^M$ for foreground objects. For each bounding box $b_j$, the Segment Anything Model (SAM)~\cite{kirillov2023segment} is employed to generate a precise segmentation mask $m_j$, resulting in object proposals represented as $(b_j, m_j)$. Each reference image and proposal is then encoded into an instance embedding. Specifically, NIDS-Net extracts average foreground features from the DINOv2 ViT patch embeddings~\cite{oquab2023dinov2}, denoted as $\mathbf{e}_i$ for reference instances and $\mathbf{e}_j$ for proposals. To enhance instance embeddings, we leverage the class token from a recent visual encoder, Perception Encoder (PE)~\cite{bolya2025perception}. A weight adapter is applied to refine these embeddings by encouraging clustering among similar instances and separation among dissimilar ones. Finally, the method matches proposal embeddings $\{\mathbf{e}_j\}$ with target embeddings $\{\mathbf{e}_i\}$ to classify each proposal, yielding predicted labels along with their corresponding bounding boxes $\{b_j\}$ and masks $\{m_j\}$. In NIDS-Net, \textit{only the weight adapter is trained using these reference images, while all other components remain frozen}.

\subsection{LLM Reasoning for Matching}

To identify the target object referred to in a natural language expression $r$ within a query image $I_Q$, we construct a candidate set $\mathcal{C} = \{(D_{c_j}, p_j, b_j, m_j)\}_{j=1}^M$, where each element consists of a retrieved object description $D_{c_j}$, a top left position $p_j=(x_j, y_j)$ of a predicted bounding box $b_j$, and its optional corresponding segmentation mask $m_j$. The object description $D_{c_j}$ is obtained by mapping the predicted instance ID $c_j$ to its associated LVLM-generated profile from the reference set. To perform matching, we prompt a Large Language Model (LLM) with $r$ and the candidate set, asking it to select the most likely match based on relative spatial position, semantic similarity and contextual cues. For simplicity, the matching process considers only the object description $D_{c_j}$ and position $p_j$, omitting the bounding box $b_j$ and mask $m_j$. This method leverages the LLM's capacity for nuanced language understanding and reasoning, enabling accurate resolution of referring expressions, even in the presence of visually similar objects or ambiguous phrasing. Our approach is interpretable and aligned with human-like comprehension, thereby improving grounding accuracy in complex scenes.

\begin{figure*}
    \centering
\includegraphics[width=0.9\linewidth]{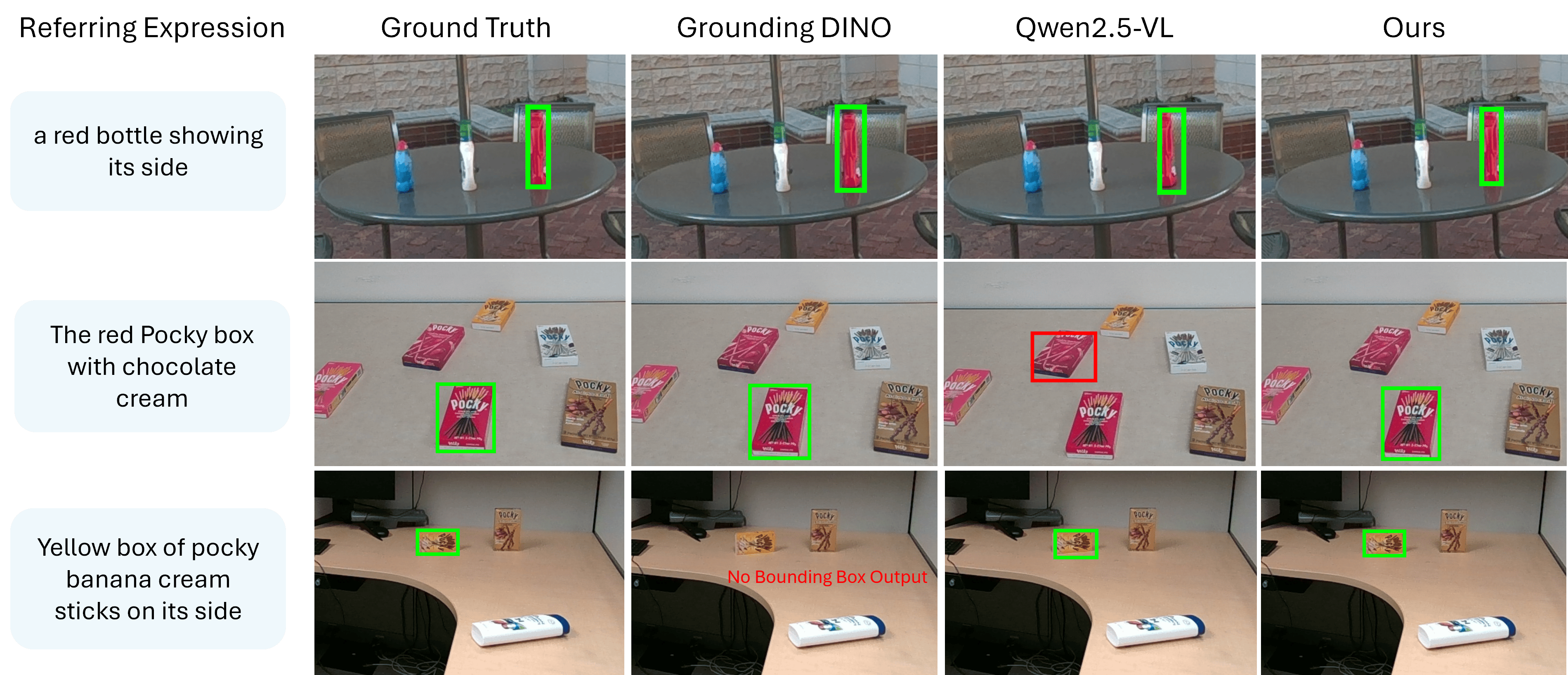}
    \caption{Visual examples of our method and other models. Green bounding boxes indicate accurate predictions, while red ones represent incorrect predictions. } 
    \label{fig:case}
\end{figure*}

\section{Experiments}

In our framework, we train only the weight adapter of NIDS-Net~\cite{lu2024adapting} using reference images. The adapter, which consists of two linear layers, is trained for 640 epochs within a few minutes. Following NIDS-Net, the training process utilizes the Adam optimizer~\cite{kingma2014adam} with a learning rate of $1 \times 10^{-3}$ and a batch size of 1024, employing the InfoNCE loss function~\cite{oord2018representation, chen2020simple}. 
All experiments are run on a server with 4 $\times$ NVIDIA A100 GPUs. The object description generation prompts for LVLMs and the matching prompts for LLMs are included in the appendix. 

\begin{table}[ht]
\centering
\caption{Visual grounding results on our MultimodalGround dataset. DINOv2~\cite{oquab2023dinov2} and PE~\cite{bolya2025perception} are different visual encoders to obtain instance embeddings in our method. }\label{tab:results}
\scalebox{0.85}{
\begin{tabular}{l|cccc}
\hline
\multicolumn{1}{c|}{Methods} & $Acc_{0.5}$ & $Acc_{0.75}$ & $Acc_{0.9}$ & $mAcc$ \\ \hline
Ferret~\cite{you2023ferret} & 0.00 & 0.00 & 0.00 & 0.00 \\
GroundingGPT~\cite{li2024groundinggpt} & 6.90 & 0.58 & 0.00 & 1.99 \\
Grounding DINO T~\cite{liu2024grounding} & 62.53 & 48.59 & 2.93 & 45.47 \\
Grounding DINO B~\cite{liu2024grounding} & 70.76 & 63.98 & 13.68 & 57.79 \\
InternVL2.5-1B~\cite{chen2024expanding} & 12.98 &  4.21& 0.35 &  6.09\\
InternVL2.5-2B~\cite{chen2024expanding} & 14.39 &  6.08& 0.47 &  7.45\\
InternVL2.5-4B~\cite{chen2024expanding} & 44.09 &  24.68& 3.51 & 26.98 \\
InternVL2.5-8B~\cite{chen2024expanding} & 55.04 & 17.68 & 1.52 &26.88  \\
Qwen-VL-Chat~\cite{DBLP:journals/corr/abs-2308-12966} & 58.83 & 40.23 &9.82  & 40.82 \\
Qwen-VL~\cite{DBLP:journals/corr/abs-2308-12966} & 65.96 &40.58  &6.32  &43.12  \\

Qwen2.5-VL-3B~\cite{bai2025qwen2} & 43.16 & 39.42 &20.00  & 37.63 \\
Qwen2.5-VL-7B~\cite{bai2025qwen2} & 75.79 & 74.50 & 47.25 &70.76  \\ 
Qwen2.5-VL-72B~\cite{bai2025qwen2}& 84.09 & 82.69 & 47.37 & 77.87  \\ 
\hline
MRVG-Net (DINOv2~\cite{oquab2023dinov2}, \textbf{Ours}) & 80.70 & 80.23 & 75.56 & 79.75 \\
MRVG-Net (PE~\cite{bolya2025perception}, \textbf{Ours}) & \textbf{86.43} & \textbf{85.96} & \textbf{80.47} & \textbf{85.38} \\
\hline
\end{tabular}
}
\end{table}

\textbf{Evaluation metrics}. Given a query image, a referring expression, and a set of reference images, a model predicts a bounding box described by the expression. To evaluate the grounding performance, we calculate the Intersection over Union (IoU) between the predicted and ground-truth bounding boxes. Predictions with an IoU greater than 0.5 are classified as true positives, and those below the threshold are false positives. This accuracy ($Acc0.5$) is computed as the number of all true positives divided by the total number of all testing samples. To comprehensively evaluate the localization capabilities of models, as suggested by~\cite{chen2024revisiting}, we calculate accuracy at different IoU thresholds: $Acc0.75$, $Acc0.9$, and $mAcc$, which calculates the average accuracy between IoU thresholds 0.5 and 0.9, in increments of 0.05.

\begin{figure}
    \centering
\includegraphics[width=0.95\linewidth]{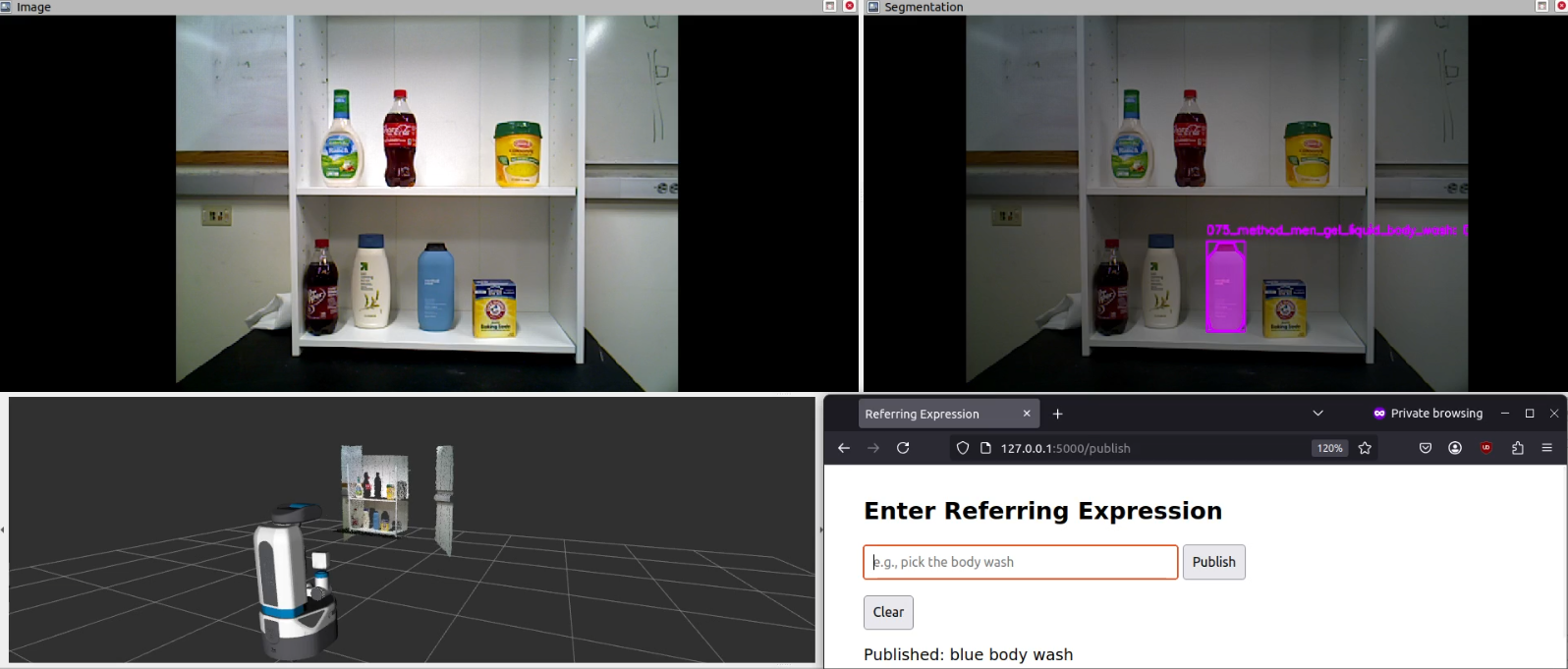}
    \caption{A real-world experiment. Given the referring expression \textit{“blue body wash”}, our method accurately and efficiently detects the target object.}
    \label{fig:test}
\end{figure}

\begin{figure*}
    \centering
\includegraphics[width=0.85\linewidth]{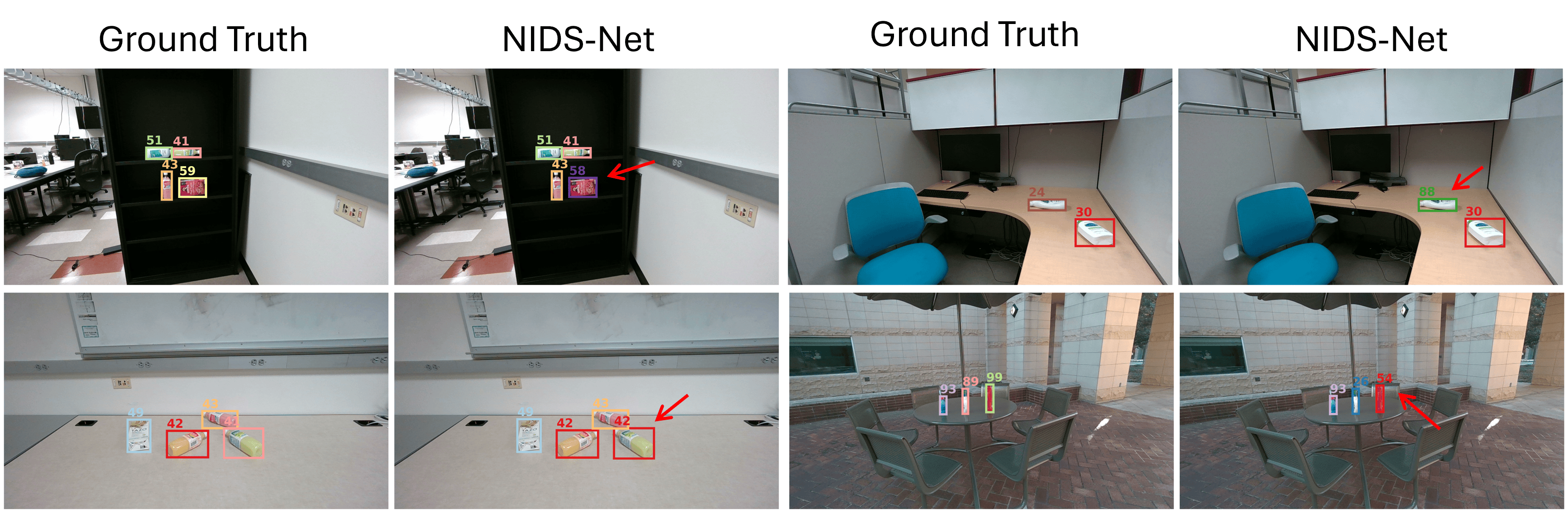}
    \caption{False detections by NIDS-Net, with red arrows indicating incorrect class labels. }
    \label{fig:failure}
    \vspace{-3mm}
\end{figure*}

\begin{table*}[ht]
\centering
\caption{Grounding results with different combinations of description generating models, matching strategies and matching models based on NIDS-Net with DINOv2. }
\label{tab:match}
\scalebox{1.0}{
\begin{tabular}{ccccccc}
\hline
Model & Matching Strategy & Matching Models & $Acc_{0.5}$ & $Acc_{0.75}$ & $Acc_{0.9}$ & $mAcc$ \\ \hline
\multirow{4}{*}{GPT-4o mini} & \multirow{2}{*}{Joint} & GPT-4o-mini & 78.01 & 77.54 & 73.45 & 77.17 \\
 &  & GPT-4o & 75.44 & 74.94 & 70.64 & 74.53 \\ \cline{2-7} 
 & \multirow{2}{*}{Independent} & GPT-4o-mini & 77.78 & 77.43 & 73.22 & 76.95 \\
 &  & GPT-4o & 80.00 & 79.53 & 74.97 & 79.05  \\ \hline
\multirow{4}{*}{GPT-4o} & \multirow{2}{*}{Joint} & GPT-4o-mini & 78.60 & 78.13 & 73.57 & 77.71 \\
 &  & GPT-4o & 79.18 & 78.71 & 74.27 & 78.26 \\ \cline{2-7} 
 & \multirow{2}{*}{Independent} & GPT-4o-mini & 78.13 & 77.78 & 73.68 & 77.31 \\
 &  & GPT-4o & \textbf{80.70} & \textbf{80.23} & \textbf{75.56} & \textbf{79.75} \\ \hline
\end{tabular}
}
\end{table*}

\subsection{Experimental Results}

\textbf{Comparison to other methods.} We display the evaluation results of different models on our MultimodalGround dataset in Table~\ref{tab:results}. For other visual grounding models, we include their prompts in the appendix~\ref{sec:baseline_prompt}. 
 While other models perform visual grounding in a zero-shot manner, our method is the only one to work in a few-shot manner by leveraging reference images for the multimodal reference visual grounding task. Our method outperforms all others across all metrics, demonstrating its superior capability in addressing a range of challenging scenarios in the visual grounding task. Furthermore, the high value of $Acc_{0.9}$ indicates that the bounding boxes of our method accurately localize the target objects. 

\textbf{Qualitative results.} We present the visual results of our dataset in Fig. \ref{fig:case}. While other models may fail to detect due to the absence of visual cues, our method leverages reference images to achieve accurate detections.

\textbf{Runtime.} For our dataset, Qwen2.5-VL-72B requires 7,969 seconds to complete inference using 8×A100 GPUs. In comparison, our method achieves the same task in 493 seconds on a single RTX A5000 (24 GB). This corresponds to a 16.2× reduction in runtime, while simultaneously delivering higher accuracy. By significantly improving both computational efficiency and performance, our approach enables robots to complete perception tasks rapidly and proceed to subsequent actions, such as cooking or bartending, without being hindered by excessive inference delays. To evaluate our method in real-world scenarios, we conducted experiments using a Fetch robot. A video in the project page demonstrating these real-world tests is provided, with an example setup shown in Fig.\ref{fig:test}.

\subsection{Ablation Study}
\label{sec:ablation_study}

\textbf{Object description models.} We evaluate different LVLMs for generating object descriptions using the same prompt. As shown in Table \ref{tab:match}, GPT-4o~\cite{hurst2024gpt} outperforms GPT-4o-mini~\cite{hurst2024gpt} when the matching strategy and matching model are the same, indicating its strong ability in the description generation task. Furthermore, upon reviewing the descriptions generated by both models, GPT-4o generally provides more detailed and informative descriptions than GPT-4o-mini.

\textbf{Matching strategies and matching models.} Two matching strategies are employed for associating a set of object descriptions with a set of referring expressions. \textbf{Joint Matching} entails simultaneously matching both sets, where each referring expression is compared to the entire set of object descriptions within a collective context. Therefore, joint matching processes all the referring expressions in a query image parallelly. In contrast, \textbf{Independent Matching} involves independently matching each referring expression with the set of object descriptions, treating each expression separately without accounting for the interactions between the referring expressions. The prompts of the two matching strategies are presented in the appendix~\ref{sec:app_matching}. Table \ref{tab:match} demonstrates that independent matching yields superior performance, as incorrect matches do not impact the accuracy of other matches. Table \ref{tab:match} also shows that GPT-4o outperforms GPT-4o-mini for the matching task in most cases on the MultimodalGround dataset, due to its superior capabilities.


\subsection{Failure Cases}

The failure cases of our method are primarily due to detection errors in NIDS-Net, as shown in Fig. \ref{fig:failure}. NIDS-Net with Perception Encoder~\cite{bolya2025perception} achieves an Average Precision (AP) of 73.3, AP at IoU threshold 0.50 of 79.7, and AP at IoU threshold 0.75 of 79.1. Incorrect class label predictions lead to mismatches between object descriptions and referring expressions. The use of a more advanced few-shot detector than NIDS-Net is expected to improve our framework's performance. 
More details of the detection performance of NIDS-Net can be found in the appendix~\ref{sec:app_detection}. 

\textbf{Failure analysis.} NIDS-Net has 228 misclassifications out of 855 predictions, while MRVG-Net experiences only 116 failures. This discrepancy arises because misclassifications from NIDS-Net do not always result in grounding failures. For instance, phrases like ``the leftmost bottle'' depend on relative spatial relationships, which can be inferred from the bounding boxes of predictions. Additionally, NIDS-Net may misclassify one instance as another similar one, where the descriptions are close enough for LLMs to still correctly locate the target object.

\section{Conclusion}


In this work, we introduced the Multimodal Reference Visual Grounding (MRVG) task, which incorporates both textual queries and multiple reference images to enhance visual grounding. Given the object descriptions from an LVLM, our proposed MRVG-Net method leverages a few-shot object detector to process reference images, followed by an LLM for matching between object descriptions and referring expressions. MRVG-Net outperforms existing models in our introduced MultimodalGround dataset, demonstrating its potential for real-world applications. This work opens new avenues for developing more robust and context-aware visual grounding systems.

\textbf{Limitations.} Few-shot detectors such as NIDS-Net can have errors in object detection. During inference, processing referring expressions sequentially can be computationally intensive and token-consuming. Therefore, our method may incur high computational costs and processing time when the query image contains a large number of items.







\section*{ACKNOWLEDGMENT}

We would like to express our sincere gratitude to the annotators who contributed to the creation of our dataset. In particular, we thank Praneeth Champooranan, Michael Ugochukwu, Itay Kadosh and Peter Driscoll for their meticulous and dedicated efforts in annotating the data. Their contributions were invaluable to the success of this work.

This work was supported in part by the DARPA
Perceptually-enabled Task Guidance (PTG) Program under contract number HR00112220005, the National Science Foundation (NSF) under Grant Nos. 2346528, 2520553 and 2513070, and the NVIDIA Academic Grant Program Award.

\bibliographystyle{IEEEtran}
\bibliography{colm2025_conference}


\onecolumn
\clearpage
\appendix
\section{Appendix}

\subsection{More Training Details}

\label{sec:app_detection}

The results of varying numbers of training epochs for the weight adapter of NIDS-Net are presented in Table \ref{tab:epoch}. We select 640 epochs as it provides stable and robust performance. AP, AP50, and AP75 are metrics used to evaluate object detection models, where AP is the average precision, AP50 is the precision at an IoU threshold of 0.5, and AP75 is the precision at an IoU threshold of 0.75. The object detection results of not using the weight adapter in NIDS-Net are presented in Table \ref{tab:WA}.

\begin{table}[ht]
\centering
\caption{Detection results with different numbers of training epochs.}\label{tab:epoch}
\begin{tabular}{ccccc}
\hline
Number of Training Epoch & 80 & 160 & 320 & 640 \\ \hline
AP & 59.9 & 58.6 & 56.3 & \textbf{60.6} \\
AP50 & 64.8 & 63.4 & 60.9 & \textbf{65.5} \\
AP75 & 64.6 & 63.1 & 60.5 & \textbf{65.2} \\ \hline
\end{tabular}
\end{table}

\begin{table}[ht]
\centering
\caption{Detection results with and without the weight adapter.}\label{tab:WA}
\begin{tabular}{ccccc}
\hline
Visual Encoder & Using Weight Adapter & AP & AP50 & AP75 \\ \hline
\multirow{2}{*}{DINOv2~\cite{oquab2023dinov2}} & $\times$ & 58.0 & 62.7 & 62.5 \\
 & $\checkmark$ & 60.6 & 65.5 & 65.2 \\ \hline
\multirow{2}{*}{Perception Encoder~\cite{bolya2025perception}} & $\times$ & 66.0 & 71.7 & 71.1 \\
 & $\checkmark$ & \textbf{73.3} & \textbf{79.7} & \textbf{79.1} \\ \hline
\end{tabular}
\end{table}





\subsection{Visual Grounding Prompts}
\label{sec:baseline_prompt}

We present the following visual grounding prompts for LVLMs. 

\begin{AIBoxBreak}{Visual Grounding Prompts}

\assistanttwomsg{Image}
\$\{image\}\$

\bigskip
\assistanttwomsg{Prompt}
Provide the bounding box coordinates of the region this sentence describes: \textless ref \textgreater \$\{ language instruction\} \$ \textless  /ref\textgreater
\end{AIBoxBreak}

\subsection{Object Description Generation}

\label{sec:app_object}

Given one representative image of an object, we show the prompt to generate the object information via a Large Vision-Language Model (LVLM). GPT-4o and GPT-4o-mini employ the same prompts. 

\begin{AIBoxBreak}{GPT-4o-mini Prompts}

\assistantonemsg{System Prompt}

You are an expert at structured data extraction. You will be given a picture. Please extract information and convert it into the given structure.\\

\assistanttwomsg{Image}

\begin{center}
    \includegraphics[width=0.6\textwidth]{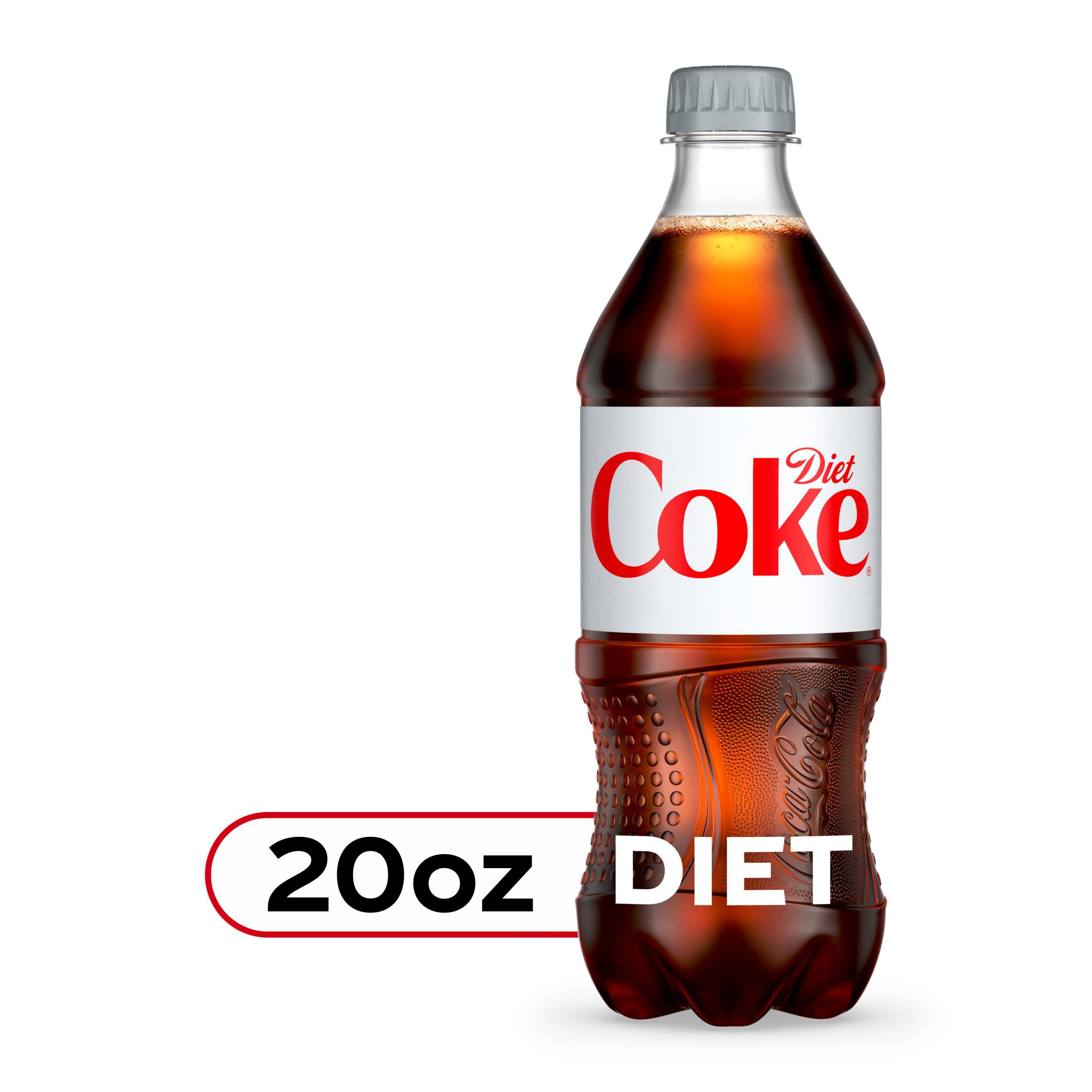}
\end{center}
\bigskip
\assistanttwomsg{Prompt}

You are given an image of an item on a flat surface (on a table, ground, etc.).
Please first carefully read and understand the image in detail.
If there are multiple items, only carefully look through one of them. 
Then, describe the item in detail by following the steps and format below.

1. Shape: Please describe the shape or type of the item, such as a bottle, 
bag, round item, square item, etc.

2. Colors: Please describe all the colors on or in the item, such as label colors, 
text colors, cover colors, etc. The item may be covered by multiple colors. 
Please describe all of them one by one. For example, bottle: transparent, liquid 
in the bottle: black, the main color of the bag: green, the text on the item: black, etc.

3. Texts: Please extract all texts on the item with the position and color of the text. 
For example, "ingredients: on the surface, black". If there is no recognized text, 
please only output "None".

4. Function: Please describe the usage of the item in the given picture.

5. Summary of the item: Please summarize the above descriptions in sentences one-by-one.

\end{AIBoxBreak}

\begin{AIBoxBreak}{GPT-4o-mini Output}

\begin{lstlisting}[language=json]
{
    "shape": "bottle",
    "colors": [
        {
            "description": "the main color of the liquid",
            "color": "brown"
        },
        {
            "description": "the color of the bottle",
            "color": "transparent"
        },
        {
            "description": "the color of the label",
            "color": "white"
        },
        {
            "description": "the color of the text on the label",
            "color": "red"
        },
        {
            "description": "the color of the cap",
            "color": "gray"
        },
        {
            "description": "the color of the text below the bottle",
            "color": "black"
        }
    ],
    "texts": [
        {
            "text": "Coke",
            "position": "on the label, red"
        },
        {
            "text": "Diet",
            "position": "on the label, red"
        },
        {
            "text": "20oz",
            "position": "below the bottle, black"
        },
        {
            "text": "DIET",
            "position": "below the bottle, black"
        }
    ],
    "name": "002_coca-cola_soda_diet_pop_bottle",
    "function": "This item is a beverage container holding Diet Coke, a low-calorie soft drink.",
    "summary": "The item is a bottle in the shape of a typical soft drink container. It features a brown liquid inside, with a transparent bottle and a white label. The label has red text that reads 'Coke' and 'Diet'. Additionally, there is black text below the bottle indicating '20oz' and 'DIET'. The cap of the bottle is gray. This bottle is used to store and serve Diet Coke."
}
\end{lstlisting}
\end{AIBoxBreak}

\begin{AIBoxBreak}{GPT-4o Prompts}

\assistantonemsg{System Prompt}

You are an expert at structured data extraction. You will be given a picture. Please extract information and convert it into the given structure.\\

\assistanttwomsg{Image}

\begin{center}
    \includegraphics[width=0.6\textwidth]{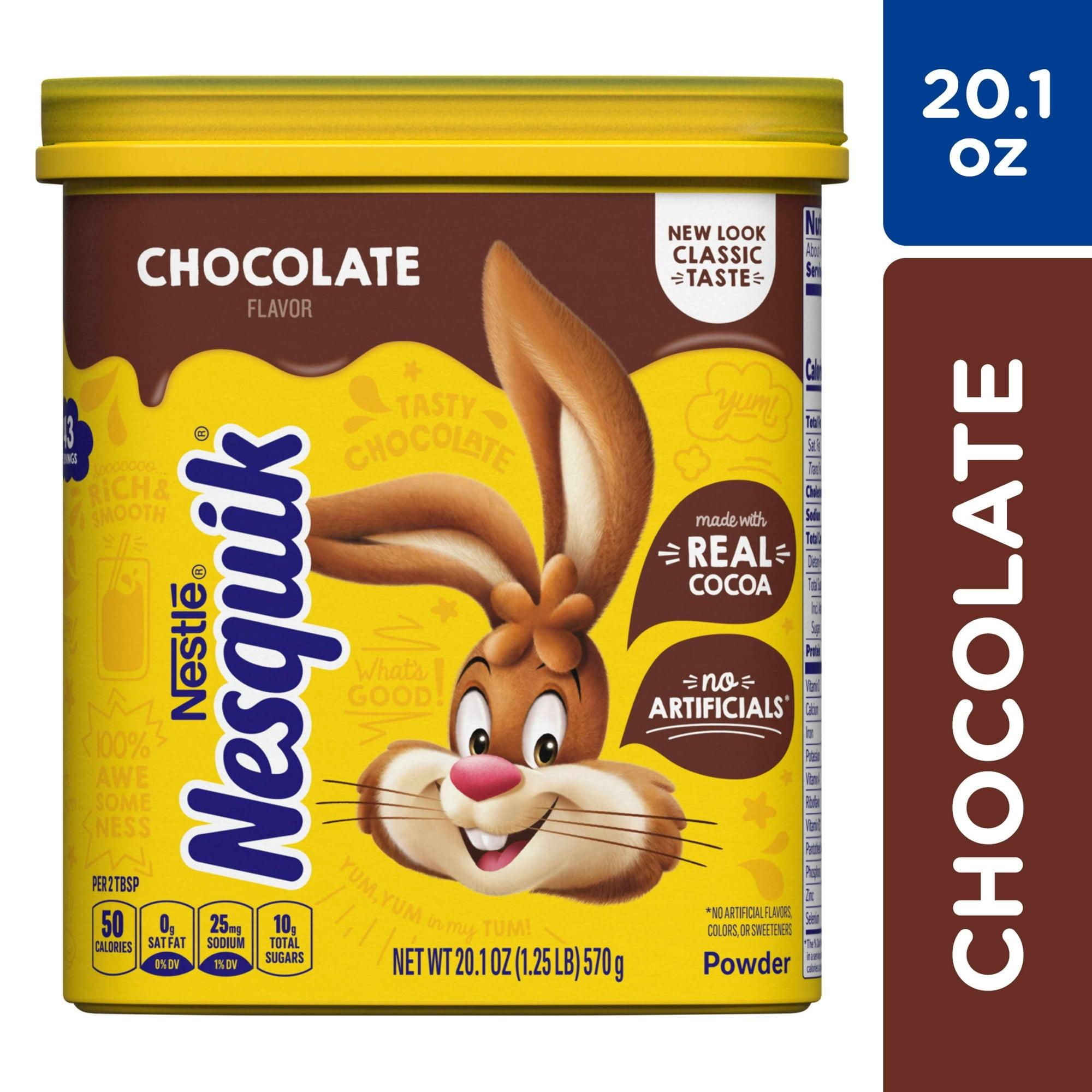}
\end{center}
\bigskip
\assistanttwomsg{Prompt}

You are given an image of an item on a flat surface (on a table, ground, etc.).
Please first carefully read and understand the image in detail.
If there are multiple items, only carefully look through one of them. 
Then, describe the item in detail by following the steps and format below.

1. Shape: Please describe the shape or type of the item, such as a bottle, 
bag, round item, square item, etc.

2. Colors: Please describe all the colors on or in the item, such as label colors, 
text colors, cover colors, etc. The item may be covered by multiple colors. 
Please describe all of them one by one. For example, bottle: transparent, liquid 
in the bottle: black, the main color of the bag: green, the text on the item: black, etc.

3. Texts: Please extract all texts on the item with the position and color of the text. 
For example, "ingredients: on the surface, black". If there is no recognized text, 
please only output "None".

4. Function: Please describe the usage of the item in the given picture.

5. Summary of the item: Please summarize the above descriptions in sentences one-by-one.

\end{AIBoxBreak}

\begin{AIBoxBreak}{GPT-4o Output}

\begin{lstlisting}[language=json]
{
   "shape":"Container",
   "colors":[
      {
         "description":"Lid color",
         "color":"Yellow"
      },
      {
         "description":"Main body color",
         "color":"Yellow"
      },
      {
         "description":"Text color",
         "color":"White, Blue, Brown"
      },
      {
         "description":"Illustration color",
         "color":"Brown, Pink, White"
      }
   ],
   "texts":[
      {
         "text":"CHOCOLATE",
         "position":"Top left",
         "color":"White"
      },
      {
         "text":"Nesquik",
         "position":"Left side",
         "color":"Blue"
      },
      {
         "text":"made with REAL COCOA",
         "position":"Right side",
         "color":"White"
      },
      {
         "text":"no ARTIFICIALS",
         "position":"Right side",
         "color":"White"
      },
      {
         "text":"50 CALORIES",
         "position":"Bottom left",
         "color":"Blue"
      },
      {
         "text":"NET WT 20.1 OZ (1.25 LB) 570 g",
         "position":"Bottom",
         "color":"Blue"
      }
   ],
   "function":"The item is a container of chocolate-flavored powder for making chocolate milk or similar beverages.",
   "summary":"The item is a container with a yellow lid and body. It features white, blue, and brown text. Illustrations are in brown, pink, and white. Text includes 'CHOCOLATE', 'Nesquik', 'made with REAL COCOA', and more. It is used for making chocolate-flavored drinks.",
   "filename":"060_nesquik_chocolate_powder"
}
\end{lstlisting}
\end{AIBoxBreak}

\subsection{LLM Matching}

\label{sec:app_matching}

For the two matching strategies, we present the following prompts to match the object descriptions of predicted objects with the referring expressions. GPT-4o and GPT-4o-mini are using the same prompts.

\subsubsection{Joint Matching}

\begin{AIBoxBreak}{GPT-4o / GPT-4o-mini}

\assistantonemsg{System Prompt}

You are an expert in information matching. Your task is to match items from a given list of descriptions to corresponding inquiries based on relevance. Each inquiry only matches one item description and appears once in the final output.\\
Each item description includes positional information, where the first value represents the x-axis (horizontal position) and the second value represents the y-axis (vertical position). A higher x-axis value indicates the item is positioned further to the right. A higher y-axis value indicates the item is positioned lower.\\
Once you determine the matches, convert them into the specified output format.\\

\assistanttwomsg{Prompt}

Items' Description:\\
Item ID: 5:
\begin{lstlisting}[language=json]
- Description: {"shape": "bottle", "colors": [{"description": "main color of the liquid", "color": "dark brown"}, {"description": "color of the bottle label", "color": "red"}, {"description": "color of the cap", "color": "burgundy"}, {"description": "color of the text on the label", "color": "white"}, {"description": "color of the establishment year", "color": "white"}], "texts": [{"text": "Dr Pepper", "position": "on the label, white"}, {"text": "Est. 1885", "position": "on the label, white"}, {"text": "23", "position": "on the label, white"}, {"text": "20 OZ", "position": "on the label, white"}], "name": "005_dr_pepper_soda_pop_bottle", "function": "This item is a beverage, specifically a carbonated soft drink, intended for consumption.", "summary": "The item is a bottle with a dark brown liquid inside, indicating a soda. The bottle label is red with white text. The cap is burgundy, and the item is identified as Dr Pepper, established in 1885. It contains 20 ounces of the beverage."}
- Position: (438, 346)
\end{lstlisting}

Item ID: 6:
\begin{lstlisting}[language=json]
- Description: {"shape": "bottle", "colors": [{"description": "main color of the bottle", "color": "orange"}, {"description": "cap color", "color": "blue"}, {"description": "label color", "color": "blue"}, {"description": "text color on the label", "color": "white"}, {"description": "text color for the size", "color": "black"}], "texts": [{"text": "FANTA", "position": "on the label, white"}, {"text": "orange", "position": "on the label, white"}, {"text": "20oz", "position": "below the bottle, black"}], "name": "006_fanta_orange_fruit_soda_pop_bottle", "function": "A carbonated soft drink flavored with orange.", "summary": "The item is a bottle in the shape of a standard soda bottle. It is predominantly orange with a blue cap and a blue label. The label features white text that reads 'FANTA' and 'orange'. Below the bottle, there is a black text indicating '20oz'. This item is a carbonated soft drink flavored with orange."}
- Position: (327, 193)
\end{lstlisting}

Item ID: 7:
\begin{lstlisting}[language=json]
- Description: {"shape": "Bottle", "colors": [{"description": "Main color of the bottle", "color": "blue"}, {"description": "Text color on the label", "color": "white"}, {"description": "Label color", "color": "black"}], "texts": [{"text": "POWER ADE", "position": "on the label, white"}, {"text": "MOUNTAIN BERRY BLAST", "position": "on the label, white"}, {"text": "50% MORE ELECTROLYTES", "position": "on the label, white"}, {"text": "20 oz", "position": "on the surface, black"}], "name": "007_powerade_mountain_berry_blast", "function": "The item is a sports drink designed to provide hydration and electrolytes, commonly used during physical activity or exercise.", "summary": "The item is shaped like a bottle. The main color of the bottle is blue, with white text on the label and a black background. The texts include 'POWER ADE', 'MOUNTAIN BERRY BLAST', and '50% MORE ELECTROLYTES' in white, and '20 oz' in black on the surface. This is a Powerade Mountain Berry Blast drink, which is used for hydration and replenishment of electrolytes."}
- Position: (650, 316)
\end{lstlisting}
\bigskip
Inquiries:\\
Inquiry ID: 1, Inquiry Content: the orange bottle.\\
Inquiry ID: 2, Inquiry Content: the middle one.\\
Inquiry ID: 3, Inquiry Content: bottle with a black cap.
\bigskip
You are given a few inquiries. Please find matched item for each inquiry and list all answers in the given format.

\end{AIBoxBreak}

\begin{AIBoxBreak}{GPT-4o / GPT-4o-mini Output}

\begin{lstlisting}[language=json]
{
  "matches": [
    {
      "inquiry_id": 1,
      "item_id": 6
    },
    {
      "inquiry_id": 2,
      "item_id": 5
    },
    {
      "inquiry_id": 3,
      "item_id": 7
    }
  ]
}
\end{lstlisting}

\end{AIBoxBreak}

\subsubsection{Independent Matching}

\begin{AIBoxBreak}{GPT-4o / GPT-4o-mini}

\assistantonemsg{System Prompt}

You are an expert in information matching. Your task is to match items from a given list of descriptions to the given inquiry based on relevance. Each inquiry only matches one item description and appears once in the final output.\\
Each item description includes positional information, where the first value represents the x-axis (horizontal position) and the second value represents the y-axis (vertical position). A higher x-axis value indicates the item is positioned further to the right. A higher y-axis value indicates the item is positioned lower.\\
Once you determine the matches, convert them into the specified output format.\\

\assistanttwomsg{Prompt}

Items' Description:\\
Item ID: 5:
\begin{lstlisting}[language=json]
- Description: {"shape": "bottle", "colors": [{"description": "main color of the liquid", "color": "dark brown"}, {"description": "color of the bottle label", "color": "red"}, {"description": "color of the cap", "color": "burgundy"}, {"description": "color of the text on the label", "color": "white"}, {"description": "color of the establishment year", "color": "white"}], "texts": [{"text": "Dr Pepper", "position": "on the label, white"}, {"text": "Est. 1885", "position": "on the label, white"}, {"text": "23", "position": "on the label, white"}, {"text": "20 OZ", "position": "on the label, white"}], "name": "005_dr_pepper_soda_pop_bottle", "function": "This item is a beverage, specifically a carbonated soft drink, intended for consumption.", "summary": "The item is a bottle with a dark brown liquid inside, indicating a soda. The bottle label is red with white text. The cap is burgundy, and the item is identified as Dr Pepper, established in 1885. It contains 20 ounces of the beverage."}
- Position: (438, 346)
\end{lstlisting}

Item ID: 6:
\begin{lstlisting}[language=json]
- Description: {"shape": "bottle", "colors": [{"description": "main color of the bottle", "color": "orange"}, {"description": "cap color", "color": "blue"}, {"description": "label color", "color": "blue"}, {"description": "text color on the label", "color": "white"}, {"description": "text color for the size", "color": "black"}], "texts": [{"text": "FANTA", "position": "on the label, white"}, {"text": "orange", "position": "on the label, white"}, {"text": "20oz", "position": "below the bottle, black"}], "name": "006_fanta_orange_fruit_soda_pop_bottle", "function": "A carbonated soft drink flavored with orange.", "summary": "The item is a bottle in the shape of a standard soda bottle. It is predominantly orange with a blue cap and a blue label. The label features white text that reads 'FANTA' and 'orange'. Below the bottle, there is a black text indicating '20oz'. This item is a carbonated soft drink flavored with orange."}
- Position: (327, 193)
\end{lstlisting}

Item ID: 7:
\begin{lstlisting}[language=json]
- Description: {"shape": "Bottle", "colors": [{"description": "Main color of the bottle", "color": "blue"}, {"description": "Text color on the label", "color": "white"}, {"description": "Label color", "color": "black"}], "texts": [{"text": "POWER ADE", "position": "on the label, white"}, {"text": "MOUNTAIN BERRY BLAST", "position": "on the label, white"}, {"text": "50% MORE ELECTROLYTES", "position": "on the label, white"}, {"text": "20 oz", "position": "on the surface, black"}], "name": "007_powerade_mountain_berry_blast", "function": "The item is a sports drink designed to provide hydration and electrolytes, commonly used during physical activity or exercise.", "summary": "The item is shaped like a bottle. The main color of the bottle is blue, with white text on the label and a black background. The texts include 'POWER ADE', 'MOUNTAIN BERRY BLAST', and '50% MORE ELECTROLYTES' in white, and '20 oz' in black on the surface. This is a Powerade Mountain Berry Blast drink, which is used for hydration and replenishment of electrolytes."}
- Position: (650, 316)
\end{lstlisting}
\bigskip
Inquiry:\\
the orange bottle.\\
\bigskip
You are given an inquiry. Please find the best matched item and output the answer in the given format.
\end{AIBoxBreak}

\begin{AIBoxBreak}{GPT-4o / GPT-4o-mini Output}

\begin{lstlisting}[language=json]
{"item_id": 7}
\end{lstlisting}

\end{AIBoxBreak}


\end{document}